# S-Net: A Scalable Convolutional Neural Network for JPEG Compression Artifact Reduction

**Bolun Zheng,[a] Rui Sun,[b] Xiang Tian,[a,c] Yaowu Chen[d,e]**
[a]Zhejiang University, Institute of Advanced Digital Technology and Instrument, No. 38 Zheda Road, Hangzhou 310027, China
[b]Sichuan University, College of Physical Science and Technology, No.24 South Section 1, Yihuan Road, Chengdu, China
[c]Zhejiang Provincial Key Laboratory for Network Multimedia Technologies
[d]The State Key Laboratory of Industrial Control Technology, Zhejiang University, China
[e]Zhejiang University Embedded System Engineering Research Center, Ministry of Education of China

**Abstract**. Recent studies have used deep residual convolutional neural networks (CNNs) for JPEG compression artifact reduction. This study proposes a scalable CNN called S-Net. Our approach effectively adjusts the network scale dynamically in a multitask system for real-time operation with little performance loss. It offers a simple and direct technique to evaluate the performance gains obtained with increasing network depth, and it is helpful for removing redundant network layers to maximize the network efficiency. We implement our architecture using the Keras framework with the TensorFlow backend on an NVIDIA K80 GPU server. We train our models on the DIV2K dataset and evaluate their performance on public benchmark datasets. To validate the generality and universality of the proposed method, we created and utilized a new dataset, called WIN143, for over-processed images evaluation. Experimental results indicate that our proposed approach outperforms other CNN-based methods and achieves state-of-the-art performance.

## 1 Introduction

Image restoration for reducing lossy compression artifacts has been well-studied, especially for the JPEG compression standard[1]. JPEG is a popular lossy image compression standard because it can achieve high compression ratio with only minimal reduction in visual quality. The JPEG compression standard divides an input image into 8×8 blocks and performs discrete cosine transform (DCT) on each block separately. The 64 DCT coefficients thus obtained are quantized based on standard quantization tables that are adjusted with different quality factors. Losses such as blackness, ringing artifacts, and blurring artifacts are mainly introduced by quantizing the DCT coefficients.

Recently, deep-neural-network-based approaches[2, 3] have been used to significantly improved the JPEG compression artifact reduction performance in terms of peak signal-to-noise ratio (PSNR). However, these methods have several limitations. First, most existing methods[2, 4, 5] focus on the construction performance of grayscale images and try to restore each channel separately when applied to color images. However, this will introduce palpable chromatic aberrations in the reconstructed image. Second, many methods[3, 4] based on the ResNet[6] architecture try to optimize the performance by increasing the number of residual blocks in networks. Although this is an effective optimization method, determining the exact depth for maximizing the network performance without traversing all depths remains challenging. Third, as found for super-resolution convolutional neural network (SRCNN)[7], CNN-based image restoration algorithms can be used for encoding, transforming, and decoding. Most existing CNN-based algorithms use only one decoder at the end of the network; alternatively, they use a stack of layers at the tail of the network as a decoder. This is called a columnar architecture. However, we believe that a columnar architecture contains too many layers between the input layer and the loss layer. Increased network depth could make it much harder for training layers around the bottom, although residual connections[8] or some excellent optimizers[9] could mitigate this problem. Therefore, if we use only some layers of the transforming part, the results could degrade considerably.

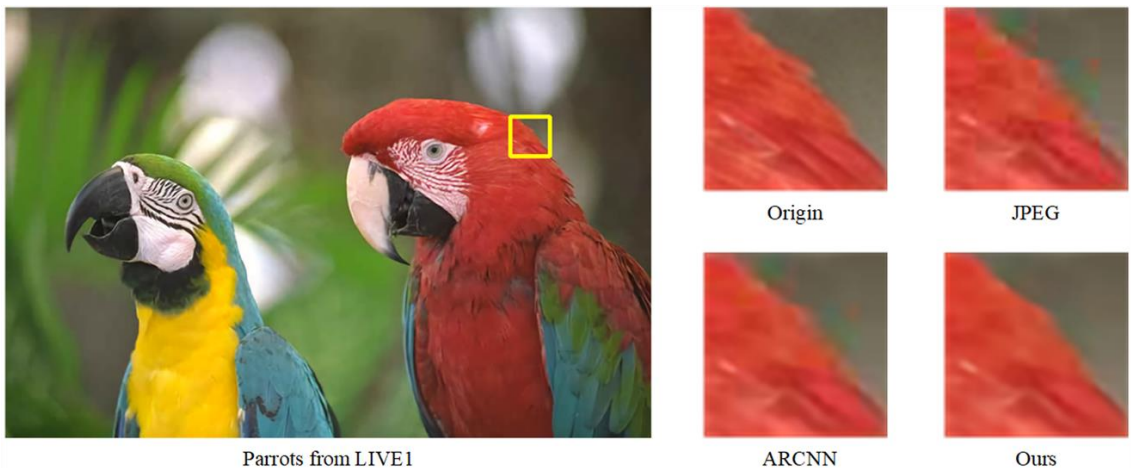

**Fig. 1** Comparison of JPEG compression (QF40) artifact reduction results of existing and proposed methods.

To solve these problems, first, we use color image pairs to train the network to restore color images directly. Second, we present a symmetric encoder-decoder model to finish the encoding and decoding tasks independently. Third, based on the greedy theory, we propose a scalable CNN called S-Net to maximize the performance of each convolutional layer in the network. We also prove that it is helpful to evaluate the influence of depth on the network performance.

We trained our models on the newly provided DIV2K[10] dataset and evaluated them on public benchmark datasets (LIVE1[11] and BSDS500[12]). The proposed architecture achieved state-of-the-art performance on all datasets in terms of PSNR and structural similarity index (SSIM).

## 2 Related Work

Deep convolutional networks (DCNs) are trained for image restoration tasks by converting an input image into a feature space and building nonlinear mappings between the features of the input and the target images. To exploit error back-propagation, groups of convolutional filters that construct DCNs are learned during the training procedure so that they can be used for convoluting an input image for a specific image restoration task. SRCNN is the first DCN-based image restoration method that has been proposed for image super-resolution (SR). It uses bicubic interpolation to up-sample the low-resolution image and train a three-layer CNN to restore the up-sampled image. Based on SRCNN, ARCNN[2, 13], a four-layer CNN, was proposed to reduce JPEG compression artifacts. However, because ARCNN does not use a pooling layer and a fully connected layer, the result deteriorated with increasing network depth, and it was difficult to guarantee convergence unless the weights of the convolutional layers were initialized carefully.

Furthermore, because DCNs show promise for high-level computer vision tasks[14, 15], many DCN-based algorithms for image restoration tasks tried to improve the network construction based on high-level computer vision algorithms. Ledig et al.[16] used a generative adversarial network (GAN)[17] to reconstruct high-resolution images by the bicubic interpolation of down-sampled low-resolution images. Li et al.[18] used GAN to solve the image dehazing problem. Other studies applied inception modules[15, 19] to image SR. Shi et al.[20] introduced a dilated convolutional layer[21] and improved the inception module in an SR network.

For image restoration tasks, very deep CNNs can only operate well with residual connections and effective optimizers. Svoboda et al.[3] used an 8-layer CNN and introduced a skip connection to learn Sobel features between a JPEG compressed image and the original image. Because JPEG compression artifacts are introduced by quantizing DCT coefficients, DDCN[4] uses the DCT domain and trains a network with both the image and the DCT domains to learn the difference between the JPEG compressed image and the original image. DDCN uses 30 residual blocks, with 10 each used for the pixel domain branch, DCT domain branch, and aggregation. Because SR has shown success with DCNs, CAS-CNN[5] imported stepped convolutional layers and deconvolutional layers (also called up-sample layers) and transformed the compression artifact reduction problem into an SR problem. Mao et al.[22] proposed RED-Net for image restoration. RED-Net uses a series of encoder-decoder pairs with symmetric skip connections to restore a noisy image. Dong et al.[23] improved RED-Net by adding a batch-normalization layer[24] and a rectified linear unit (ReLU)[25] layer to the output of each convolutional layer and deconvolutional layer[26, 27] to learn the intrinsic representations and successfully solved the image restoration and image classification problems using the same pre-trained networks. However, these two approaches based on a convolutional autoencoder did not provide an explicit presentation of the relationship between the whole network and each encoder-decoder pair in it. By contrast, symmetrically connected convolutional layer pairs were more likely to be deformed residual structures rather than encoder-decoder pairs. Lim et al.[8] proposed a very deep network with 32 residual blocks for single-scale SR (EDSR) and an even deeper network with 80 residual blocks for multiscale SR (MDSR).

In general, DCN-based methods for image restoration tasks like compression artifact reduction (AR) and SR focus on improving performance by increasing the number of parameters in the network. However, larger-scale networks always incur higher computational costs and longer training procedures that may sometimes be unaffordable. Higher costs are also incurred to evaluate whether the best performance is achieved by the network. This study focuses on maximizing the network performance by using fewer parameters and minimizing the training procedure time.

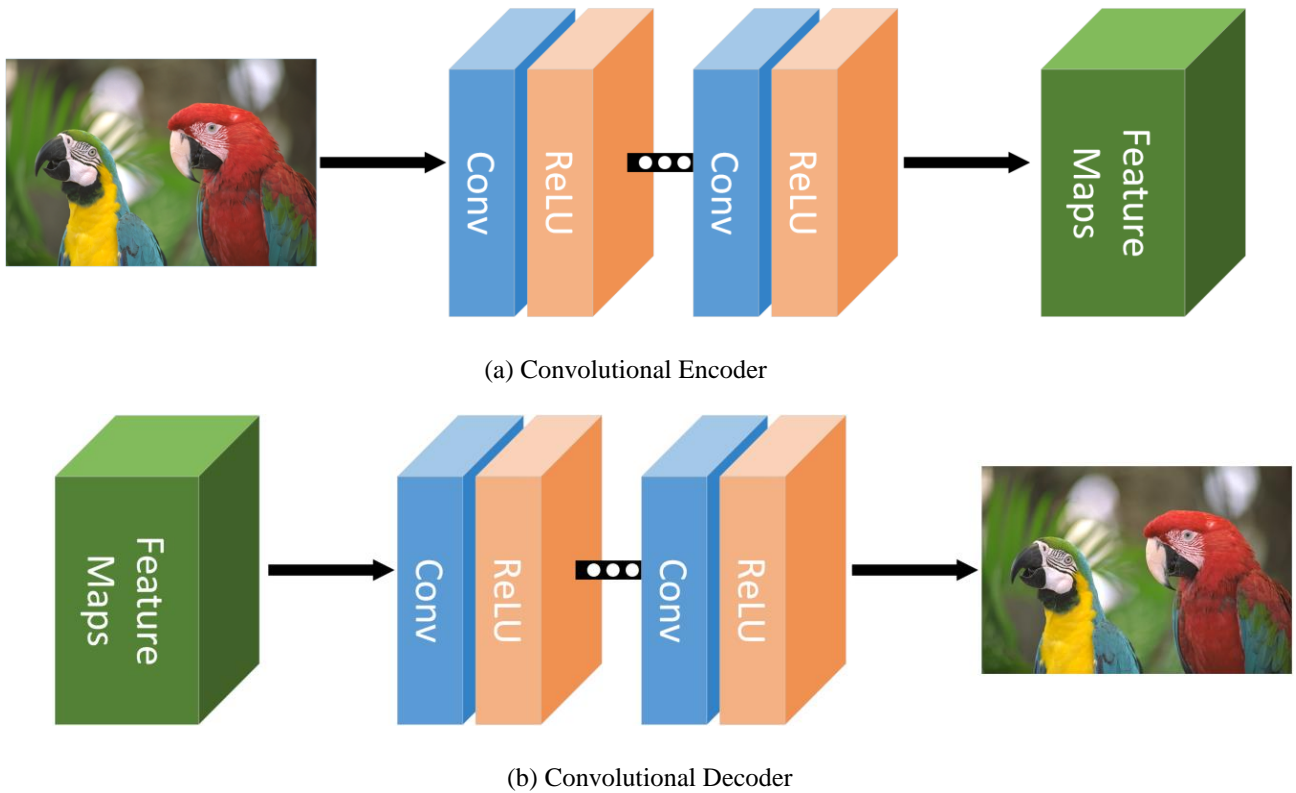

(a) Convolutional Encoder

(b) Convolutional Decoder

**Fig. 2** Constructions of convolutional encoder-decoder model.

## 3    Proposed Method

This section describes the proposed network architecture and implementation. First, we propose a convolutional encoder-decoder architecture to extract features from an input image and recover the input image from the feature domain. Then, we introduce the characterization of the greedy loss architecture for building a scalable CNN. Finally, we discuss the implementation of our proposed architecture.

*3.1 Symmetric Convolutional Encoder-Decoder Model*

A traditional autoencoder constructed by a multilayer full-connection network is a classical unsupervised machine learning algorithm for dimension reduction and feature extraction. Some SR reconstruction algorithms[28, 29] use an autoencoder to learn the sparse representation of image patches and to try to refine image patches in the sparse representation domain. Considering the relative position information of pixels in image patches, we modified the autoencoder by replacing fully connected layers with convolutional layers and built a convolutional encoder-decoder architecture.

Figure 2 shows the construction of our convolutional encoder-decoder model. The encoder and decoder blocks are formulated by a series of combinations of a convolutional layer and an activation layer. Let X be the input; the encoder and decoder blocks are expressed as:

$$\phi_i(X) = \begin{cases} \max(0, W_i * X + B_i) & i = 0 \\ \max(0, W_i * \phi_{i-1}(X) + B_i) & i > 0 \end{cases} \quad (1)$$

where $W_i$ and $B_i$ respectively represent the parameters of the $i^{\text{th}}$ convolutional filters and bias, and $*$ is the convolution operation.

The two blocks have the same number of layers and a symmetric convolutional kernel size. For example, if the encoder block contains $M$ layers and the convolutional kernel size of the $k^{\text{th}}$ ( $k \in \{1,2,3...M\}$ ) layer is $s_k \times s_k$, then the convolutional kernel size of the $k^{th}$ layer in the decoder block should be $s_{M-k+1} \times s_{M-k+1}$.

Moreover, unlike in the case of the encoder block, we tried to formulate a decoder block with transposed convolutional layers[30] instead of convolutional layers. However, although the relationship between convolution and transposed convolution seems like a key-lock relationship, this change does not result in any improvement in the datasets in our benchmark.

*3.2 Greedy Loss Architecture*

Although increasing the network depth is a simple way to improve performance, a deeper network does not always result in better performance. It is difficult to ensure the appropriate depth that maximizes the network performance without testing various depths. To solve this problem, we propose a greedy loss architecture to maximize the performance of each convolutional unit in the network. We use the encoder-decoder architecture to translate inputs from the image domain to the feature domain and to ensure that each output of a convolutional unit is limited to a fixed feature domain. We connect the decoder and loss layer after the output of each convolutional unit and hope that each unit can map the JEPG compressed features to the original features.

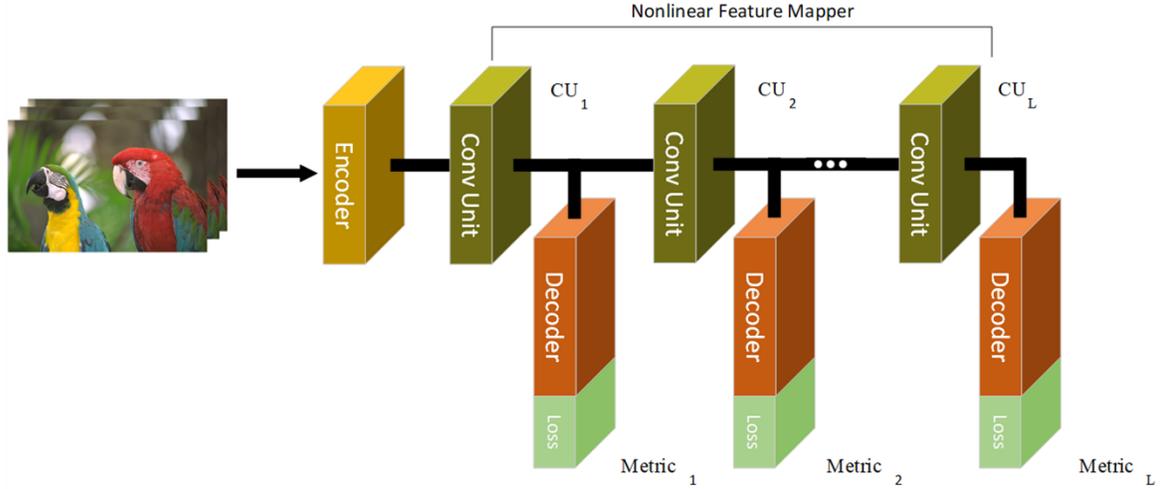

**Fig. 3** Overview of proposed network.

Convolutional units at different depths clearly receive different gradients from back-propagation; specifically, shallower ones receive more gradients than deeper ones, and this could be helpful for optimizing the performance of shallower layers. On the other hand, the greedy loss architecture constrains the output of each convolutional unit; this is conductive for avoiding gradient explosion[19] when training large-scale convolutional networks. Furthermore, it is feasible to determine the network performance with different depths without training the whole network repeatedly by analyzing the loss from each convolutional unit with fixed depth. Section 4 discusses the experimental evaluation of the network performance.

**Table 1** Construction of convolutional encoder-decoder model.

| Conv Layer | Filter Size | |
|---|---|---|
| | Encoder | Decoder |
| 1 | 5×5×256 | 3×3×256 |
| 2 | 3×3×256 | |
| | | 5×5×3 |

*3.3 Building a Scalable Convolutional Neural Network*

*3.3.1 Overview*

Our network architecture can be divided into three parts: encoder, decoder and nonlinear feature mapper. The encoder uses JPEG compressed image patches as inputs and translates them to the feature domain. A nonlinear feature mapper learns a nonlinear functions to map features from anamorphic image patches to uncompressed image features. The decoder translates mapped features back to the image domain. Figure 3 shows an overview of our network architecture. Unlike other deep neural networks in which all parts have to operate in combination in order for the network to function, these three parts construct a scalable convolutional neural network in which the operation of even one part enables the network to function normally. For example, if part of the nonlinear feature mapper is removed, the network can give a comparatively good result. Moreover, even if the nonlinear feature mapper is completely removed, the symmetrically encoder-decoder pairs can still give quite improved outputs.

*3.3.2 Architecture*

The encoder and decoder are both formulated using two convolutional layers with a ReLU activation layer. Table 1 lists the construction of the encoder and decoder. Because the second convolutional layer of the decoder is connected to the loss layer, its number of channels depends on the channel size of the input/output images. A nonlinear feature mapper is formulated by a series of shortcut connections. Because these shortcut connections in low-level image processing using DCNs are always constructed using only convolutional layers, we call these full-convolutional shortcut connections as convolutional units. We denote these convolutional units as $CU_1$, $CU_2$, …, $CU_L$, where $L$ is the total number of convolutional units. All these convolutional units have the same structure, and they are formulated by convolutional layers and activation layers.

There are active discussions about the problem of deeper networks or wider networks[31, 32]. The creators of ResNet preferred deeper networks and tried to make the network as thin as possible in order for it to have only a few parameters. Some recently proposed DCN-based methods for image restoration adopted this strategy to construct their networks. However, the wide ResNet (WRN)[33] was developed based on the rationale that deeper networks can lead to less gradient flowing and fewer convolutional units can result in useful intern representations being learned. Further, the performance also suffers significantly from this very deep structure when the depth of the network is dynamically scaled. Motivated by this observation, we designed our S-Net with larger width and shallower depth. We set $L = 8$, and all convolutional layers in the convolutional units comprise filters with a kernel size of 3×3 and a channel size of 256.

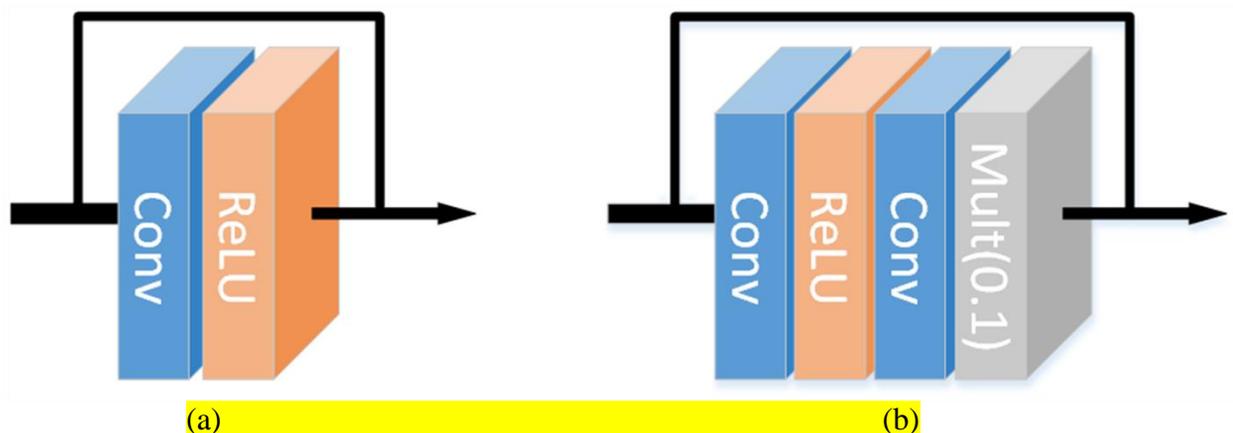

(a)                                                                       (b)

**Fig. 4** Structures of different convolutional units: (a) classic residual structure and (b) advanced residual structure.

*3.3.3 Convolutional Unit*

Residual connections have been widely and successfully used in many image restoration algorithms[16, 34, 35]. A convolutional unit constructed with residual connection is the basic component of a network, and it plays an important role in the network performance. Here, we compare two widely used convolutional units: classic residual structure, the simplest residual connection structure in which the residual branch is constructed using a convolutional layer and a ReLU activation layer, and advanced residual structure, first proposed by Peng et al.[36] for boundary refining in image segmentation and which shows great performance for image restoration tasks[8, 35]. Although the batch-normalization layer is a basic component in the residual connection structure, it has been found that removing them from the network can improve the network performance for image SR tasks[8]. We found that this modification is also effective for AR tasks and therefore applied this modification to our network. Figure 4 shows the configuration of both structures. The residual branch of the advanced residual structure contains two convolutional layers and a ReLU activation layer. Table 2 lists the parameters of our networks constructed using these two different convolutional units.

Several researchers replaced the ReLU layer in networks with a parametric rectified linear unit (PReLU)[2, 5, 37] layer. PReLU imports a learnable parameter $\alpha$ to restrict the negative output, whereas ReLU compulsively cuts the negative output to zero.

$$PR(x) = \begin{cases} x & x \geq 0 \\ \alpha x & x < 0 \end{cases} \quad (2)$$

We tried using PReLU layers to replace ReLU layers in our network. However, doing so did not result in any improvement; instead, it increased the computational and memory costs. Therefore, we use only ReLU as the activation function in our following experiments.

**Table 2** Size of parameters of proposed architectures.

| Architecture | Number of Parameters | | Total | |
|---|---|---|---|---|
| | Classic | Advanced | Classic | Advanced |
| Encoder | 0.58M | 0.58M | - | - |
| Decoder | 0.58M | 0.58M | - | - |
| $CU_1$ | 0.56M | 1.12M | 1.72M | 2.29M |
| $CU_2$ | 0.56M | 1.12M | 2.29M | 3.41M |
| $CU_3$ | 0.56M | 1.12M | 2.85M | 4.54M |
| $CU_4$ | 0.56M | 1.12M | 3.41M | 5.66M |
| $CU_5$ | 0.56M | 1.12M | 3.97M | 6.78M |
| $CU_6$ | 0.56M | 1.12M | 4.54M | 7.91M |
| $CU_7$ | 0.56M | 1.12M | 5.10M | 9.04M |
| $CU_8$ | 0.56M | 1.12M | 5.66M | 10.16M |

*3.3.4 Training*

PSNR is the most universal evaluation indicator for image restoration tasks. It can be represented as follows:

$$PSNR=10\log_{10}(MSE(\hat{X},Y)) \quad (3)$$

where $Y$ is the target image; $\hat{X}$, the restored image; and MSE, the mean square error. To maximize the PSNR of the reconstructed image, we use the MSE loss as the loss function of each metric in the greedy loss architecture. The loss weight of each metric is equated to others, and the sum of the weights of all metrics is one. The final training loss function is

$$\tau(\Theta) = \frac{1}{MN}\sum_{i=1}^{N}\sum_{j=1}^{M}\left\|\kappa_j(X^i;\Theta)-Y^i\right\| \quad (4)$$

where $N$ is the number of samples in a training batch; $M$, the number of metrics; and $\kappa_j$, the output of the $j^{th}$ metric in the greedy loss architecture.

We used the adaptive moment estimation (Adam)[9] method as the optimizer during the training procedure. Adam is a recently proposed optimization algorithm that has been proved to be effective for training very deep networks. We used the default parameters (beta_1 = 0.9, beta_2 = 0.999, epsilon = $10^{-8}$)[9] as specified in the original paper.

## 4 Experimental Evaluation

*4.1 Dataset*

The DIV2K[10] dataset has been recently proposed in the NTIRE2017 challenge for single-image SR[10]. The dataset consists of 1000 2K-resolution images, of which 800 are training images, 100 are validation images, and the remaining 100 are testing images. Because the testing dataset was prepared for image SR and the ground truth has not been not released, we could not compare performances using this dataset. Instead, we evaluated the performance of our proposed method and compared it with other state-of-the-art methods on other known datasets.

LIVE1[11] and BSDS500[12] are two known datasets that are frequently used to validate the performance of proposed approaches in reducing JPEG compressed artifacts. LIVE1 is a publicly released dataset that contains 29 images for image quality assessment. BSDS500 is a publicly released database for image segmentation that contains 200 training images, 100 validation images, and 200 test images. Quantitative evaluations are conducted on the 29 images in the LIVE1 dataset and the 200 test images in the BSDS500 dataset.

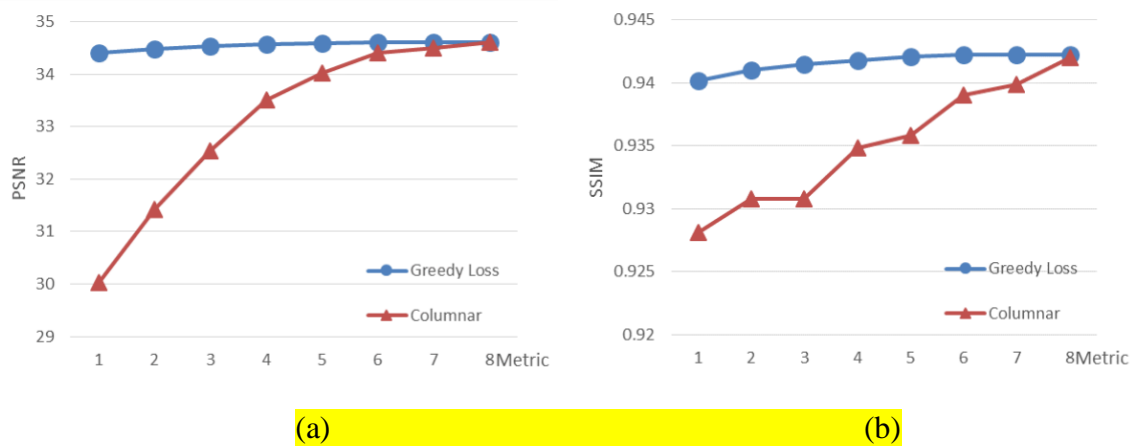

(a)                  (b)

**Fig. 5** Performance comparison of models with convolutional unit of advanced residual structure trained using greedy loss architecture and columnar architecture on LIVE1 dataset with JPEG quality of 40: (a) PSNR and (b) SSIM.

*4.2 Training*

For training, we extracted 48×48 RGB image patches from training images in the DIV2K dataset with random steps from 37 to 62 as the input. Although the JPEG compression algorithm is applied to each 8×8 patch, taking a random step to avoid integral multiples of eight can significantly enhance the network performance, as in the case of DDCN[4]. The initial learning rate was set to $10^{-4}$ at the start of the training procedure, and subsequently halved after every set of $10^4$ batch updates until it was below $10^{-6}$. All network models for different convolutional units were trained with $2\times10^5$ batch updates. Considering the limitation of computational resources, the batch size was set to 16. We first trained models with JPEG quality of 40 (QF40) to evaluate the performance of different convolutional units. Then, we fine-tuned the pre-trained QF40 models for JPEG qualities of 10 (QF10) and 20 (QF20) with initial learning rate of $1\times10^{-5}$ and $4\times10^4$ batch updates. During fine-tuning, the learning rate was also halved after every set of $10^4$ batch updates until it reached $10^{-6}$ or below. We performed the experiments using the Keras framework with a TensorFlow backend on an NVIDIA K80 GPU server. Training of the QF40 model took five days and two days was required to fine-tune the QF20 and QF10 models.

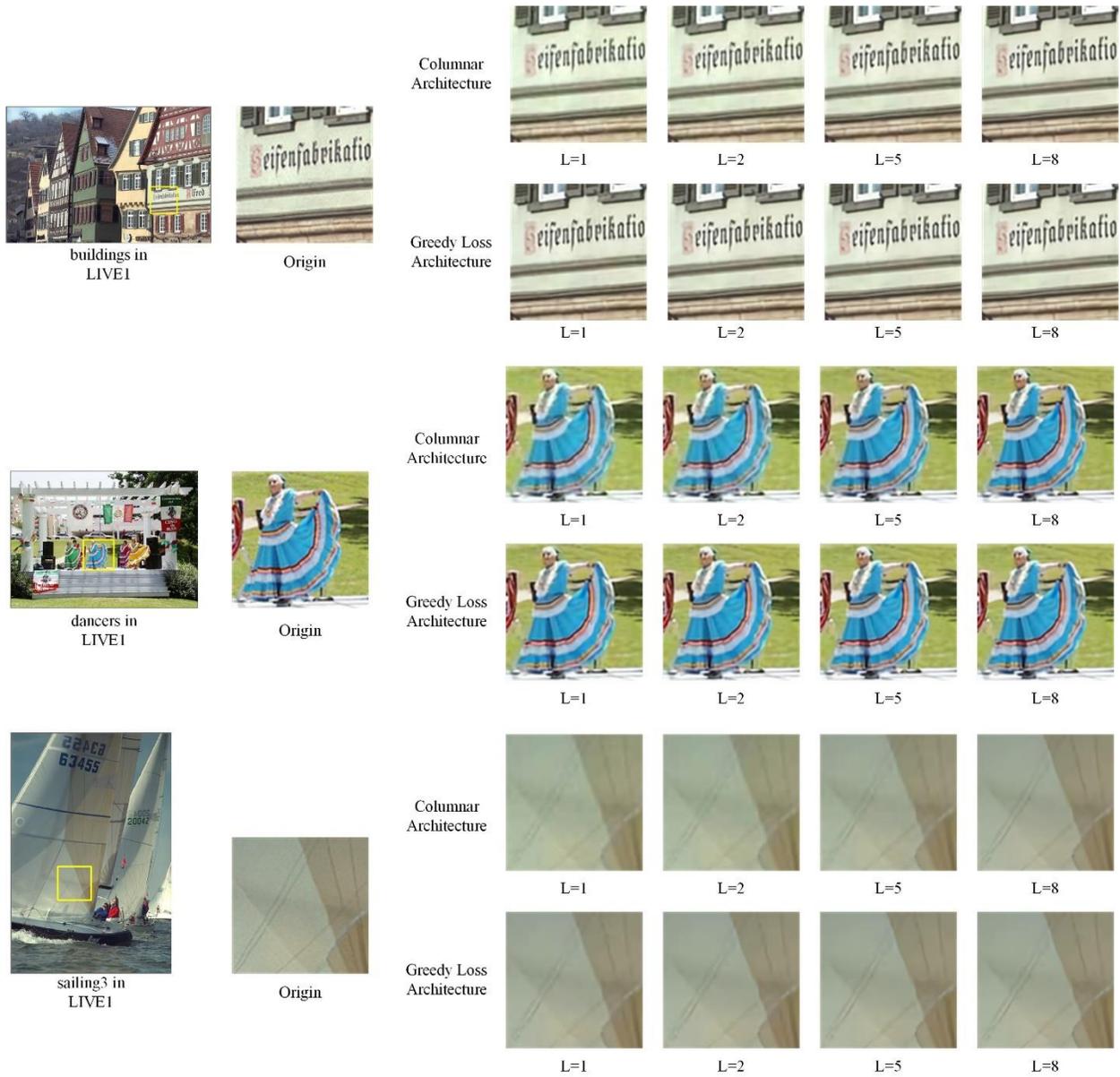

**Fig. 6** Comparison of details of images reconstructed by models with convolutional unit of advanced residual structure trained using greedy loss architecture and columnar architecture on LIVE1 dataset (JPEG quality = 40).

*4.3 Greedy Loss Architecture Performance Evaluation*

We measured the PSNR and SSIM with only the y-channel considered, and used standard MATLAB library functions for the evaluations. We trained the network models with the convolutional unit of the advanced residual structure using the columnar architecture and greedy loss architecture for the JPEG quality of 40. For fairness, these two models were trained with the same image patches and same learning rate during the whole training procedure. Figure 5 shows the results in terms of PSNR and SSIM for the LIVE1 dataset. We compared the performances of the two network models after $2\times10^5$ batch updates on the LIVE1 dataset. Although the greedy approach may lead to a local optimum, the model trained using the greedy loss architecture showed better performance in terms of both PSNR and SSIM compared with the model trained using the columnar architecture. Furthermore, the proposed model showed better performance for intermediate outputs than the columnar architecture. Figure 6 shows the reconstructed images. The greedy loss architecture significantly improved the consistency of color accuracy and texture sharpness in different metrics. With some convolutional units removed from the network, the result obtained using the greedy loss architecture was obviously more stable than that obtained using the columnar architecture.

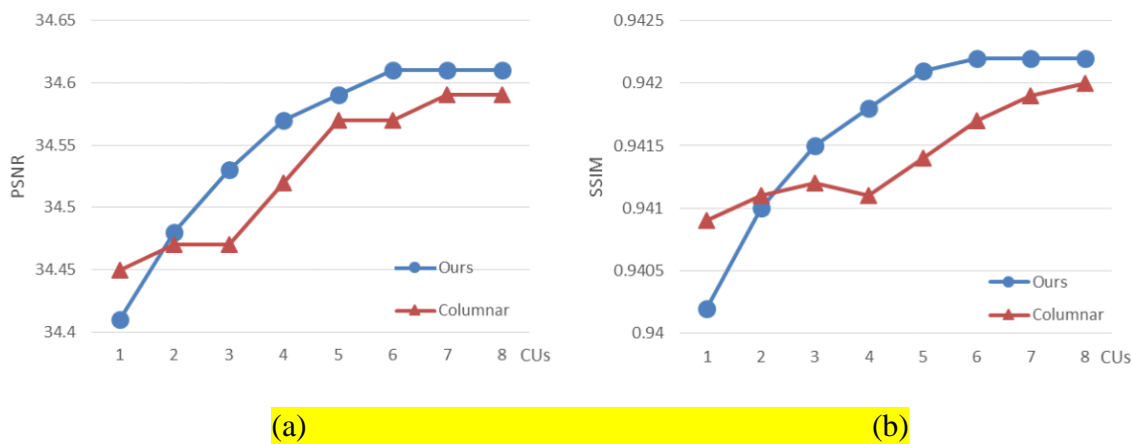

(a)  (b)

**Fig. 7** Comparison of the performance of metrics in S-Net using columnar architecture CNNs with same network scale on LIVE1 dataset with convolutional unit of advanced residual block (JPEG quality = 40): (a) PSNR and (b) SSIM.

We also compared S-Net with columnar architecture CNNs with 1–8 convolutional units. Figure 7 shows the obtained results. The performance of metric two in S-Net is very close to that of the columnar architecture CNN with two convolutional units. The subsequent metrics in our model all show better performance than the columnar architecture CNNs for the same network scale. These results also show that scalable CNN is effective for evaluating the performance improvement with increased network depth. The results for columnar architecture CNNs show that the performance of the nonlinear feature mapper with four convolutional units is close to eight, and the decrease in PSNRs from eight to four is less than 0.1 dB. Furthermore, the performance did not significantly improve for more than six convolutional units. The results of the metrics in S-Net are identical. This indicates that the greedy loss architecture in S-Net is effective for evaluating the improvement achieved with increased network scale.

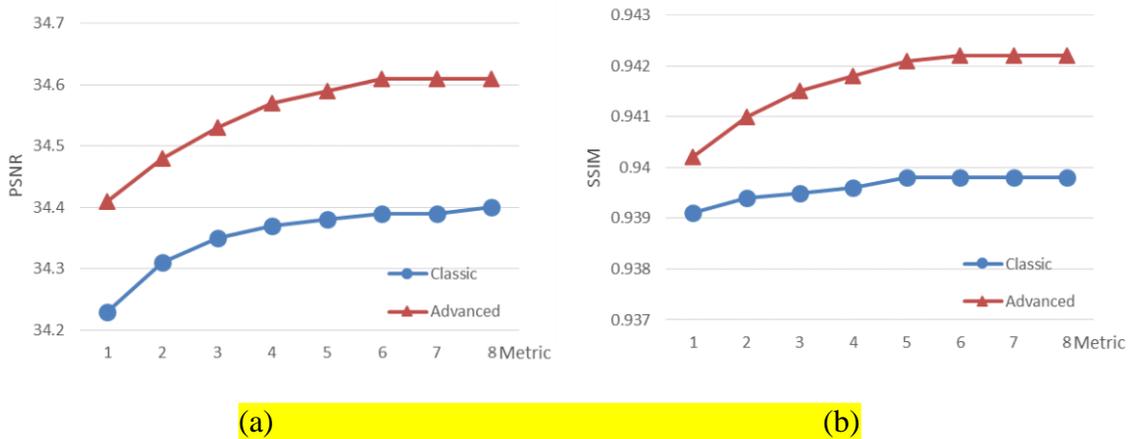

(a) (b)

**Fig. 8** Comparison of performance of models with different convolutional units on LIVE1 dataset (JPEG quality = 40): (a) PSNR and (b) SSIM.

*4.4 Evaluation of Different Convolutional Units*

Figure 8 shows quantitative evaluation results of network models with different convolutional units on the LIVE1 dataset. The model trained with an advanced residual block showed significant improvement compared to that of the model trained with a classic residual block. The advanced residual block model provides average enhancement of 0.20 dB for each metric with the trade-off of double the number of parameters in the nonlinear feature mapper.

*4.5 Comparisons to State-of-the-Art*

We compared our models with the state-of-the-art models ARCNN[2], DDCN[4], and CAS-CNN[5] on the BSDS500 and LIVE1 datasets. Because only ARCNN provides open source code and a pre-trained model, the results for ARCNN were obtained from experiments conducted by ourselves, while the results for DDCN and CAS-CNN were obtained from the reports presented in corresponding papers. Further, no qualitative comparison could be carried out for DDCN and CAS-CNN. As shown in Table 3, our model constructed with convolutional units of the advanced residual structure showed significant improvements compared to the other methods for all public benchmark datasets. Figure 9 shows some qualitative results for the BSDS500 dataset.

**Table 3** Comparison of our approaches with existing methods on public benchmark datasets. Boldface indicates the best performance.

| Dataset | JPEG Quality | Average PSNR (dB)/SSIM | | | | | | |
|---|---|---|---|---|---|---|---|---|
| | | JPEG | ARCNN | CAS-CNN | DDCN | Metric1 (proposed) | Metric2 (proposed) | Metric8 (proposed) |
| LIVE1 | 40 | 32.93/0.9255 | 33.63/0.9306 | 34.10/0.937 | -/- | 34.41/0.9402 | 34.48/0.9410 | **34.61/0.9422** |
| | 20 | 30.62/0.8816 | 31.40/0.8886 | 31.70/0.895 | -/- | 32.05/0.9034 | 32.13/0.9046 | **32.26/0.9067** |
| | 10 | 28.36/0.8116 | 29.13/0.8232 | 29.44/0.833 | -/- | 29.67/0.8415 | 29.75/0.8435 | **29.87/0.8467** |
| BSDS500 | 40 | 32.89/0.9257 | 33.55/0.9296 | -/- | 34.27/0.9389 | 34.27/0.9394 | 34.33/0.9401 | **34.45/0.9413** |
| | 20 | 30.61/0.8811 | 31.28/0.8854 | -/- | 31.88/0.8996 | 31.97/0.9017 | 32.04/0.9028 | **32.15/0.9047** |
| | 10 | 28.39/0.8098 | 29.10/0.8198 | -/- | 29.59/0.8381 | 29.64/0.8391 | 29.71/0.8410 | **29.82/0.8440** |

We also compared the computational efficiency of the proposed method with those of other state-of-the-art methods. All algorithms were implemented on a K80 GPU server with a single GPU core. We measured the computational efficiency in terms of million color pixels per second (MCP/s). However, because there is a significant difference between the qualities of the images reconstructed by ARCNN and other state-of-the-art methods, ARCNN is not included in this comparison although it is quite fast. Figure 10 shows the computational efficiencies. S-Net with one and two convolutional units shows improved image quality and computational efficiency.

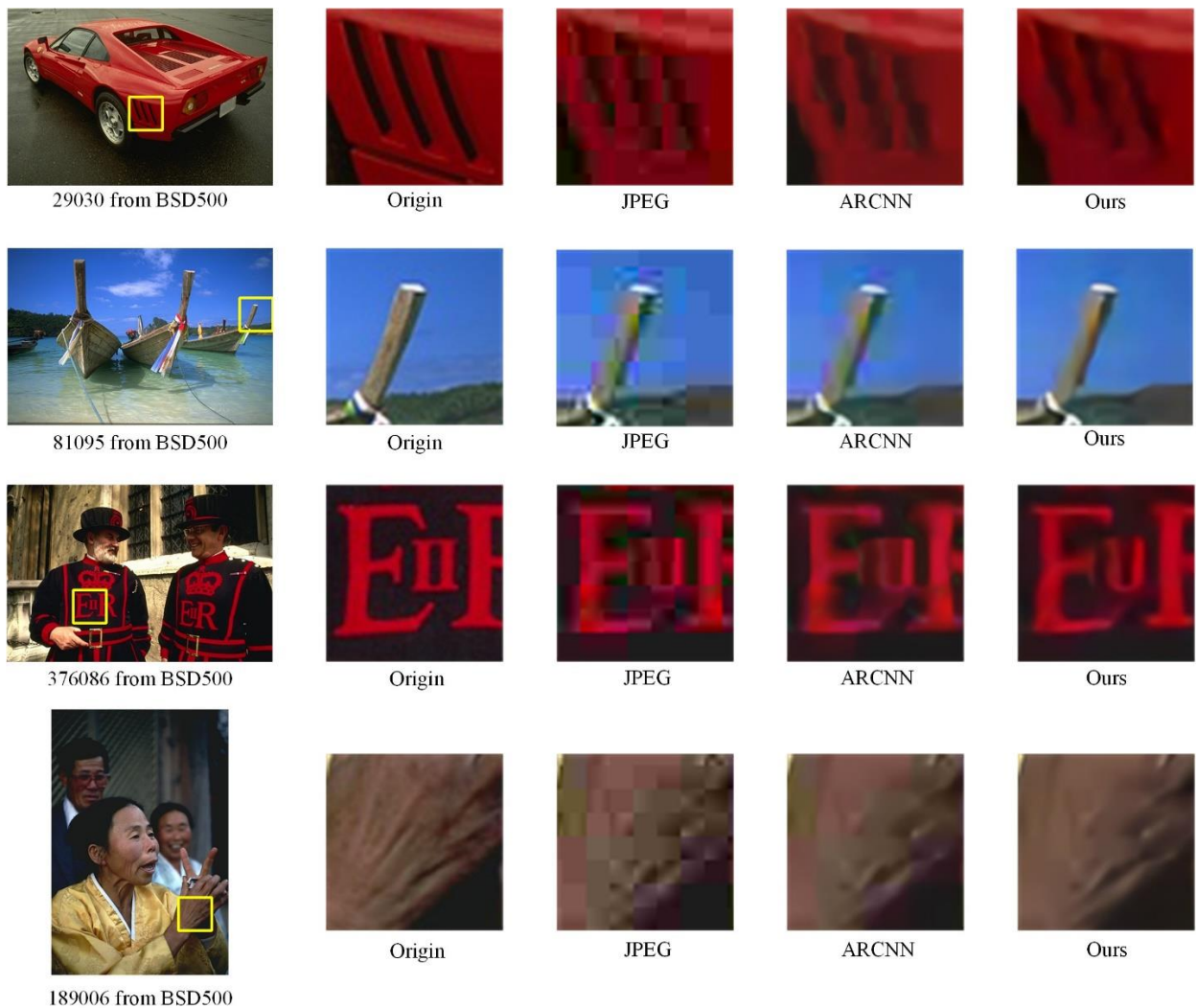

**Fig. 9** Comparison of our model with state-of-the-art methods for QF10 JPEG compression artifact reduction.

*4.6 Extensional Performance Evaluation on the WIN143 Dataset*

However, we noticed that the images in both LIVE1 and BSDS500 are typically everyday scenes. Further, few shooting skills or post processing technologies were utilized when getting these images. We call images acquired like this normal images. However, because we believe that these limitations may not thoroughly show the generality and universality of algorithms, as a supplement, we created an extensional dataset, called WIN143, to evaluate the algorithm performance on specially acquired or post processed images. The WIN143 dataset contains 143 desktop wallpapers with a resolution of 1920×1080 that are always used in the Windows 10 operating system. The images in WIN143 were collected from the internet and are specially shot or carefully post processed, or even generated by computer graphics technologies. Here, we call images such as these over-processed images. Compared to daily shot images, the over-processed images always get higher contrast and saturation, and their complexity and unexpected changes are obviously enhanced.

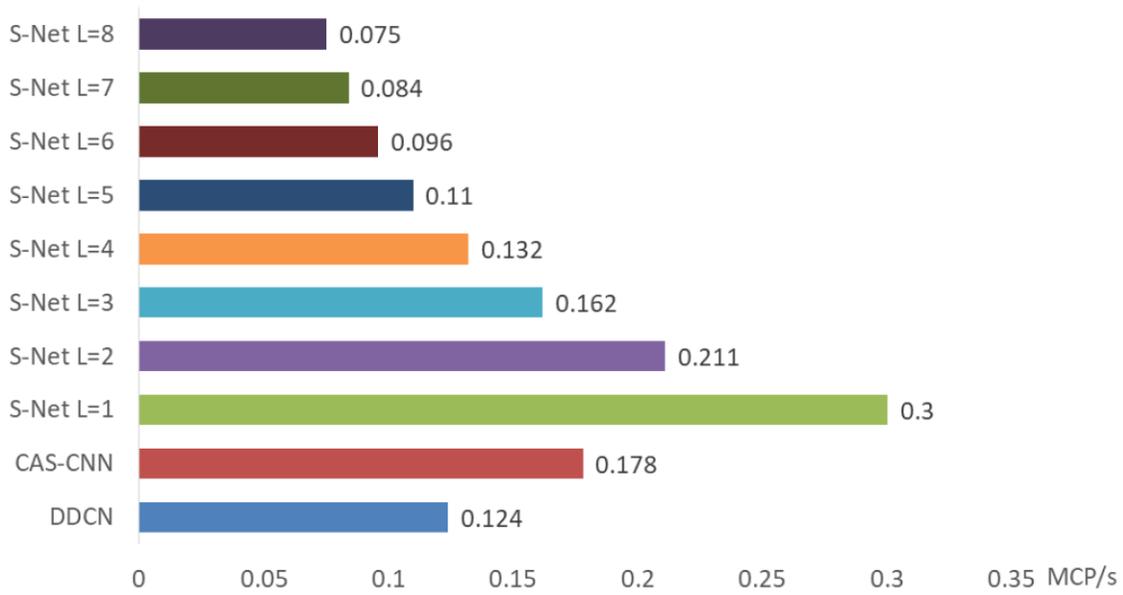

**Fig. 10** Comparison of computational efficiency of our model with those of other state-of-the-art methods.

The extensional experiment on the WIN143 dataset was conducted with a JPEG quality of 20. Because the original resolution of the images in the WIN143 dataset was too large for them to be placed in memory, we reduced the image height and width by half using the bicubic interpolation algorithm. The performance comparison results in terms of PSNR and SSIM are shown in Table 4. The performance difference between ARCNN and S-Net is significantly magnified that 28 of 143 images restored by ARCNN get even worse results in terms of PSNR while S-Net remains good performances as before. Figure 11 shows the qualitative results for the WIN143 dataset. The images restored by S-Net have higher color and intensity accuracy, especially in the dark and the smooth areas. Further, S-Net is better able to distinguish the true textures and fake textures created by JPEG compression.

Table 4 Comparison of our approaches with existing methods on the WIN143 datasets. Boldface indicates the best performance.

| Items | Average PSNR (dB)/SSIM | | | | |
|---|---|---|---|---|---|
| | JPEG | ARCNN | Mertic1 (proposed) | Metric2 (proposed) | Metric8 (proposed) |
| WIN143 | 32.95/0.9033 | 33.09/0.9106 | 34.38/0.9220 | 34.47/0.9232 | **34.61/0.9250** |
| public benchmark improvement | - | +0.69/+0.0046 | +1.37/+0.0195 | +1.44/+0.0219 | **+1.55/+0.0238** |
| WIN143 improvement | - | +0.14/+0.0073 | +1.43/+0.0187 | +1.52/+0.0199 | **+1.66/+0.0217** |

\* Here, public benchmark refers to the combination of BSDS500 and LIVE1 datasets; improvement refers to the PSNR and SSIM improvements compared to JPEG compressed images.

## 5 Conclusion

This study investigated the effects of increased network depth on network performance and proposed a scalable CNN called S-Net for JPEG compression artifact reduction. By applying a symmetric convolutional encoder-decoder model and a greedy loss architecture, S-Net dynamically adjusts the network depth. We proved that this greedy theory-based architecture does not sink into a local optimum and achieves better results than a specifically trained network under the same conditions. Furthermore, the proposed architecture is also helpful for discovering the minimal network scale with maximum possible network performance. With the greedy loss architecture, the evaluation results for the depth of the network were quickly obtained after training once, whereas several training sessions had to be applied with the conventional architecture. We compared our approach with other state-of-the-art algorithms on public benchmark datasets and achieved top ranking. We also created an over-processed image dataset, called WIN143, using images obtained from the internet. The results of extensional performance evaluation on the WIN143 dataset successfully validated the generality and universality of the proposed algorithm.

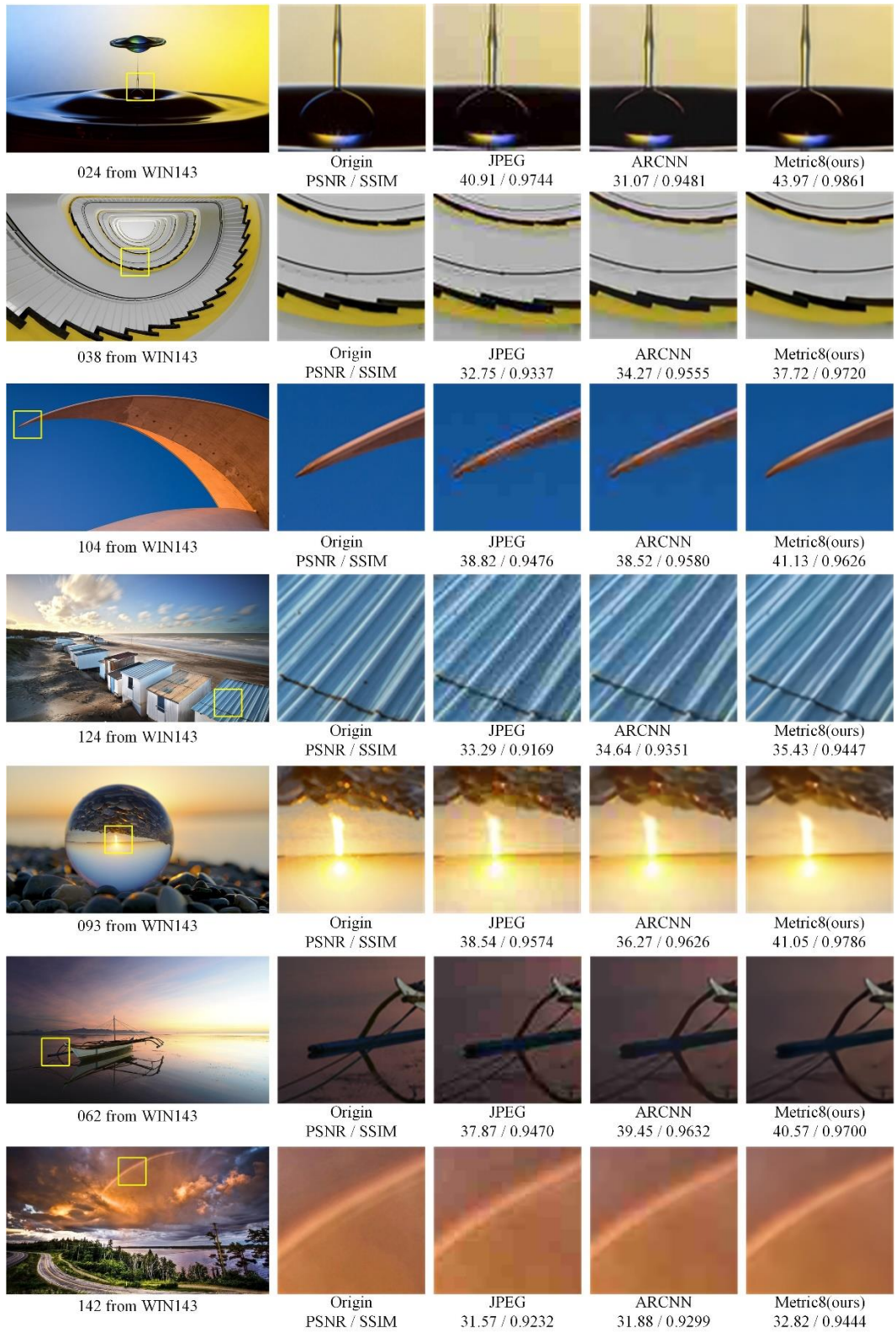

Fig. 11 Comparison of our model with state-of-the-art methods for QF20 JPEG compression artifact reduction on the WIN143 dataset.

*Acknowledgements*

This work was supported by Fundamental Research Funds for the Central Universities, China. The authors would like to thank the anonymous reviewers for their valuable comments that have helped to significantly improve this manuscript.
*References*

[1] G. K. Wallace, "The JPEG still picture compression standard," *Communications of the Acm,* 38(1), xviii-xxxiv (1991) [doi:10.1145/103085.103089]
[2] K. Yu, C. Dong, C. C. Loy *et al.*, "Deep convolution networks for compression artifacts reduction," *arXiv preprint arXiv:1608.02778* (2016)
[3] P. Svoboda, M. Hradis, D. Barina *et al.*, "Compression Artifacts Removal Using Convolutional Neural Networks," *Journal of Wscg,* 24(2), 63-72 (2016)
[4] J. Guo, and H. Chao, "Building Dual-Domain Representations for Compression Artifacts Reduction." *European Conference on Computer Vision. Springer, Cham*, 628-644 (2016)
[5] L. Cavigelli, P. Hager, and L. Benini, "CAS-CNN: A deep convolutional neural network for image compression artifact suppression." *International Joint Conference on Neural Networks. IEEE*, 752-759 (2017) [doi:10.1109/ijcnn.2017.7965927]
[6] K. He, X. Zhang, S. Ren *et al.*, "Deep Residual Learning for Image Recognition." *IEEE Conference on Computer Vision and Pattern Recognition. IEEE*, 770-778 (2016) [doi:10.1109/cvpr.2016.90]
[7] C. Dong, C. C. Loy, K. He *et al.*, "Learning a deep convolutional network for image super-resolution." *Computer Vision - ECCV 2014. Springer International Publishing*, 184-199 (2014)
[8] B. Lim, S. Son, H. Kim *et al.*, "Enhanced Deep Residual Networks for Single Image Super-Resolution." *Computer Vision and Pattern Recognition Workshops. IEEE*, 1132-1140 (2017) [doi:10.1109/CVPRW.2017.151]
[9] D. Kingma, and J. Ba, "Adam: A Method for Stochastic Optimization," *Computer Science*, (2014)
[10] E. Agustsson, and R. Timofte, "NTIRE 2017 Challenge on Single Image Super-Resolution: Dataset and Study." *Computer Vision and Pattern Recognition Workshops. IEEE*, 1110-1121 (2017) [doi:10.1109/CVPRW.2017.150]
[11] Z. Wang, A. C. Bovik, H. R. Sheikh *et al.*, "Image quality assessment: from error visibility to structural similarity," *IEEE Transactions on Image Processing,* 13(4), 600-612 (2004) [doi:10.1109/TIP.2003.819861]
[12] P. Arbelaez, M. Maire, C. Fowlkes *et al.*, "Contour Detection and Hierarchical Image Segmentation," *IEEE Transactions on Pattern Analysis & machine Intelligence,* 33(5), 898 (2011) [doi:10.1109/TPAMI.2010.161]
[13] C. Dong, Y. Deng, C. L. Chen *et al.*, "Compression Artifacts Reduction by a Deep Convolutional Network." *IEEE Conference on Computer Vision. IEEE*, 576-584 (2016) [doi:10.1109/iccv.2015.73]
24

**Bolun Zheng** is a PhD candidate of Zhejiang University and received his BSc degree from Zhejiang University, Hangzhou, China, in 2014. His current research interests include image restoration, image processing and computer vision.

**Rui Sun** is an undergraduate student of Sichuan University in Chengdu, China in 2015. She is currently participate in a project of FPGA watermark image acceleration processing.

**Xiang Tian** received his BSc and PhD degrees from Zhejiang University, Hangzhou, China, in 2001 and 2007, respectively. Currently, he is an associate professor in the Institute of Advanced Digital Technologies and Instrumentation, Zhejiang University. His major research interests include VLSI-based high-performance computing, networking multimedia systems, and parallel processing

**Yaowu Chen** received his PhD from Zhejiang University, Hangzhou, China, in 1998. Currently, he is a professor and the director of the Institute of Advanced Digital Technologies and Instrumentation, Zhejiang University. His major research interests include embedded systems, networking multimedia systems, and electronic instrumentation systems.

**Caption List**

**Fig. 1** Comparison of JPEG compression (QF40) artifact reduction results of existing and proposed methods.

**Fig. 2** Constructions of convolutional encoder-decoder model.

**Fig. 3** Overview of proposed network.

**Fig. 4** Structures of different convolutional units: (a) classic residual structure and (b) advanced residual structure.

**Fig. 5** Performance comparison of models with convolutional unit of advanced residual structure trained using greedy loss architecture and columnar architecture on LIVE1 dataset with JPEG quality of 40.

**Fig. 6** Comparison of details of images reconstructed by models with convolutional unit of advanced residual structure trained using greedy loss architecture and columnar architecture on LIVE1 dataset (JPEG quality = 40) : (a) PSNR and (b) SSIM.

**Fig. 7** Comparison of the performance of metrics in S-Net using columnar architecture CNNs with same network scale on LIVE1 dataset with convolutional unit of advanced residual block (JPEG quality = 40) : (a) PSNR and (b) SSIM.

**Fig. 8** Comparison of performance of models with different convolutional units on LIVE1 dataset (JPEG quality = 40) : (a) PSNR and (b) SSIM.

**Fig. 9** Comparison of our model with state-of-the-art methods for QF10 JPEG compression artifact reduction.

**Fig. 10** Comparison of computational efficiency of our model with those of other state-of-the-art methods.

**Fig. 11** Comparison of our model with state-of-the-art methods for QF20 JPEG compression artifact reduction on the WIN143 dataset.

**Table 1** Construction of convolutional encoder-decoder model.

**Table 2** Size of parameters of proposed architectures.

**Table 3** Comparison of our approaches with existing methods on public benchmark datasets. Boldface indicates the best performance.

**Table 4** Comparison of our approaches with existing methods on the WIN143 datasets. Boldface indicates the best performance.